\documentclass[11pt]{article}

\usepackage[a4paper, margin=1in]{geometry}
\usepackage{times}
\usepackage[T1]{fontenc}
\usepackage[utf8]{inputenc}
\usepackage{amsmath,amssymb}
\usepackage{graphicx}
\usepackage{booktabs}
\usepackage{caption}
\usepackage{subcaption}
\usepackage{hyperref}
\usepackage{xcolor}
\usepackage{authblk}
\usepackage[numbers,sort&compress]{natbib}
\usepackage{microtype}
\usepackage{enumitem}

\hypersetup{
    colorlinks=true,
    linkcolor=blue,
    citecolor=blue,
    urlcolor=blue,
    pdftitle={Deep Learning-Based Sign Language Recognition from Videos and Cross-Lingual Translation to Indian Vernaculars},
    pdfauthor={Ramesh Nandipalli, Chandranath Adak}
}

\title{\textbf{Deep Learning-Based Sign Language Recognition from Videos and Cross-Lingual Translation to Indian Vernaculars}}

\author[1]{Ramesh Nandipalli}
\author[1]{Chandranath Adak}
\affil[1]{Department of Computer Science and Engineering, Indian Institute of Technology Patna, Bihar, India}

\date{}

\begin{document}

\maketitle

\begin{abstract}
Sign language is a primary mode of communication for the global deaf and hard-of-hearing community, yet automated tools that recognize sign gestures from video and translate them into natural-language text remain limited, particularly for low-resource Indian languages. We present a two-stage deep learning pipeline that (i) classifies short sign-language video clips into English word labels using a fine-tuned VideoMAE video transformer, and (ii) translates the predicted English label into Hindi, Telugu, and Bengali using Meta AI's No Language Left Behind (NLLB-200) multilingual translation model. The classification model is fine-tuned on a 13-class subset of the AI4Bharat Indian Sign Language video corpus from IIT Madras, processing 16-frame clips sampled uniformly from each video at $224\times224$ resolution. Under a small-scale academic setting (13 classes, 197 clips, 80/20 split), the fine-tuned model reaches 99\% training accuracy and 78\% validation accuracy after 15 epochs. We provide a per-class breakdown via a confusion matrix and classification report, identify the dominant failure modes (confusable adjective pairs such as \emph{ugly}/\emph{deaf}/\emph{blind} and \emph{hat}/\emph{dress}), and describe a Streamlit-based inference demo that takes a user-uploaded video and returns the predicted English label alongside its Hindi, Telugu, and Bengali translations. We discuss the scope, limitations (small label set, isolated-word rather than continuous signing, single-signer style sensitivity, ambiguity of single-word machine translation), and directions for future work, including expanding to sentence-level generation and a larger vocabulary. Code is released to support reproducibility.
\end{abstract}


\section{Introduction}
\label{sec:intro}

Sign language is the primary mode of communication for hundreds of millions of deaf and hard-of-hearing people worldwide. Despite its importance, automated systems that can reliably recognize sign gestures and translate them into written or spoken text remain underdeveloped relative to other areas of natural language and vision research. Most deployed solutions either restrict themselves to static hand-shape classification from images, or handle only a narrow gesture vocabulary, and very few support translation into low-resource regional languages. This gap is especially pronounced in India, where dozens of major regional languages coexist alongside Indian Sign Language (ISL), and where parallel sign-to-text corpora for individual regional languages are scarce.

This work originated as an image-based classification exercise: a Vision Transformer (ViT) was trained on a custom dataset of static sign images labeled with English words and translated into Telugu, reaching 93\% training and 90\% testing accuracy. Motivated by the practical limitations of static-image recognition for real signing (which is inherently a temporal, whole-body activity), we extended this work to video. We pose the overall task as a two-stage pipeline: \emph{video-to-English} classification using a pretrained video transformer, followed by \emph{English-to-regional-language} translation using a multilingual neural machine translation model. This decomposition sidesteps the absence of large parallel sign-to-regional-language corpora by routing through English as an intermediate, high-resource pivot language.

\subsection{Problem Definition}
We address two coupled sub-problems: (1) given a short video clip of a signed word or short phrase, predict the corresponding English word label; and (2) given that English label, produce natural translations into Hindi, Telugu, and Bengali. We focus on whole-video, whole-gesture classification (rather than per-frame hand-keypoint classification), since hand-only approaches often miss facial expression and upper-body cues that are linguistically meaningful in many sign languages.

\subsection{Objectives}
The objectives of this study are to: (i) fine-tune a video transformer (VideoMAE) for isolated-word sign language video classification; (ii) build an English-to-Indian-language translation stage using a pretrained multilingual NMT model (NLLB-200); (iii) evaluate the combined pipeline on a curated subset of a large public ISL video corpus; and (iv) package the pipeline into an interactive demo suitable for non-technical end users.

\subsection{Scope and Limitations}
\label{sec:scope}
Due to compute and storage constraints of a free-tier cloud notebook environment, training was restricted to a 13-class subset (adjectives such as \emph{loud}, \emph{quiet}, \emph{happy}, \emph{sad}, \emph{beautiful}, \emph{ugly}, \emph{deaf}, \emph{blind}, and clothing nouns \emph{hat}, \emph{dress}, \emph{suit}, \emph{skirt}, \emph{shirt}) drawn from a much larger ($\sim$56\,GB) sign-video archive. The system performs isolated-word classification against a fixed label set; it does not generate free-form English sentences from continuous signing, and it is designed for offline batch inference rather than real-time streaming. These constraints are discussed further in Section~\ref{sec:limitations}.

\section{Related Work}
\label{sec:related}

\subsection{Early Approaches to Sign Language Recognition}
Early sign language recognition (SLR) systems relied on classical machine learning with hand-engineered features—Support Vector Machines, $k$-Nearest Neighbours, and Hidden Markov Models—operating on hand shape, orientation, and trajectory features extracted from instrumented gloves, depth sensors, or video frames~\citep{starner1998,cooper2011}. These approaches demonstrated feasibility but generalized poorly across signers and recording conditions, and scaled badly to large vocabularies due to the cost of manual feature design.

\subsection{Deep Learning for SLR}
The availability of larger datasets and GPU compute shifted SLR toward deep learning. Convolutional Neural Networks (CNNs) proved effective for static, image-based sign classification by learning hierarchical spatial features automatically. For dynamic, video-based signing, hybrid CNN–RNN architectures (commonly CNN feature extractors feeding LSTM or GRU temporal models) captured spatio-temporal dynamics more directly~\citep{koller2016,camgoz2018}, but such recurrent pipelines often struggled with long-range temporal dependencies and lacked an explicit mechanism for attending to the most informative frames in a clip.

\subsection{Transformer Models for Video Understanding}
The Transformer architecture~\citep{vaswani2017attention}, originally proposed for sequence modeling in NLP, was adapted to vision through the Vision Transformer (ViT)~\citep{dosovitskiy2021vit}, which treats an image as a sequence of patch tokens and applies self-attention across them. Extending self-attention to video, VideoMAE~\citep{tong2022videomae} uses a masked-autoencoding pretraining objective with an extremely high masking ratio over space-time tubes, forcing the model to learn rich spatio-temporal representations from unlabeled video. Pretrained VideoMAE encoders have since become a strong, general-purpose backbone for downstream video classification tasks, including action recognition, and are a natural candidate for sign gesture recognition given their capacity to model global dependencies across both space and time.

\subsection{Multilingual Machine Translation for Low-Resource Languages}
Translating recognized signs into natural language is typically framed as recognition followed by translation. While high-resource pairs such as English–French are well served by neural machine translation, most Indian regional languages remain comparatively low-resource, with limited parallel sign-language corpora. Meta AI's No Language Left Behind (NLLB) project~\citep{nllb2022} introduced a single multilingual model covering 200 languages, including many low-resource ones, trained with large-scale data mining and sparse mixture-of-experts architectures to substantially improve translation quality for under-served languages. NLLB models provide a practical, off-the-shelf bridge from English to Hindi, Telugu, and Bengali without requiring sign-language-specific parallel corpora in each target language.

\subsection{Positioning of This Work}
Building on these threads, we combine a pretrained VideoMAE encoder, fine-tuned for isolated-word ISL video classification, with a pretrained NLLB-200 model for English-to-regional-language translation. This English-pivot design is a pragmatic response to the scarcity of direct sign-to-regional-language parallel data in the Indian context, and is consistent with the broader SLR literature's framing of recognition and translation as a two-stage process~\citep{camgoz2020slt}.

\section{Methodology}
\label{sec:method}

\subsection{Dataset}
\label{sec:dataset}
We use a subset of the AI4Bharat Indian Sign Language video corpus released by IIT Madras~\citep{ai4bharat_isl}, which in full comprises approximately 56\,GB of labeled gesture videos organized into folders by word/label. Because of the compute and storage limits of the free-tier notebook environment used for this study, we selected 13 labels totaling approximately 2\,GB: eight adjectives (\emph{loud}, \emph{quiet}, \emph{happy}, \emph{sad}, \emph{beautiful}, \emph{ugly}, \emph{deaf}, \emph{blind}) and five clothing-related nouns (\emph{hat}, \emph{dress}, \emph{suit}, \emph{skirt}, \emph{shirt}). Each class folder contains multiple \texttt{.MOV} clips of different signers performing the same sign. We additionally surveyed, but did not train on, the Mendeley ISLR dataset and a Zenodo-hosted ISL corpus as alternative sources for future scale-up.

\subsection{Model Architecture}
\label{sec:architecture}
\begin{figure}[t]
    \centering
    \includegraphics[width=0.95\textwidth]{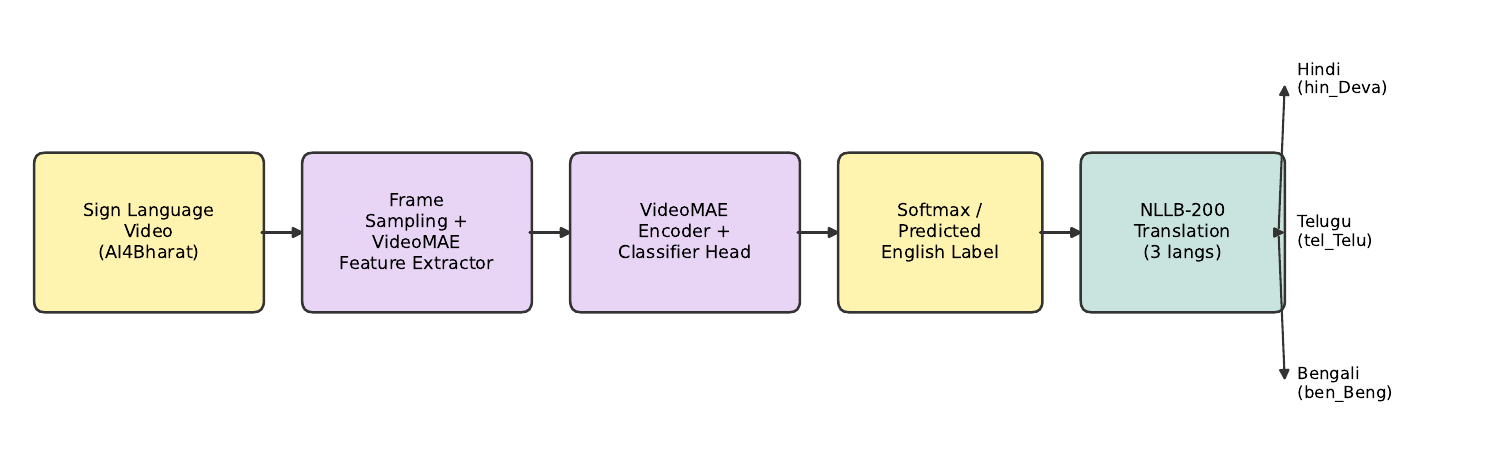}
    \caption{End-to-end pipeline: sign-language video is sampled to 16 frames, encoded by a fine-tuned VideoMAE classifier into an English label, and the label is independently translated into Hindi, Telugu, and Bengali using NLLB-200.}
    \label{fig:pipeline}
\end{figure}

Figure~\ref{fig:pipeline} summarizes the end-to-end pipeline. Each video is decoded and 16 frames are sampled uniformly across its duration; frames are resized to $224\times224$ and normalized with ImageNet statistics. The resulting clip tensor is passed to a pretrained \texttt{VideoMAEForVideoClassification} model (Hugging Face checkpoint \texttt{MCG-NJU/videomae-base}), which tokenizes the clip into space-time patches via a 3D convolutional patch embedding and processes them with a 12-layer Transformer encoder. The original ImageNet/Kinetics classification head is replaced with a new linear layer with 13 output units, one per sign class in our subset. At inference time, softmax is applied to the logits, the arg-max class is taken as the predicted English label, and the label (after stripping the leading numeric prefix used for file organization, e.g., \texttt{"37.Hat"} $\to$ \texttt{"Hat"}) is passed to the translation stage.

The translation stage uses \texttt{facebook/nllb-200-distilled-600M}~\citep{nllb2022} from Hugging Face. The predicted English label is translated independently into Hindi (\texttt{hin\_Deva}), Telugu (\texttt{tel\_Telu}), and Bengali (\texttt{ben\_Beng}) by setting the tokenizer's source language to \texttt{eng\_Latn} and forcing the decoder's beginning-of-sequence token to each target language code in turn.

\subsection{Training and Evaluation}
\label{sec:training}
The 13-class video set was split 80/20 into training and validation partitions using stratified sampling on the label, yielding 157 training clips and 40 validation clips. The classification head and the full VideoMAE encoder were fine-tuned jointly using the AdamW optimizer (learning rate $1\times10^{-5}$), cross-entropy loss, batch size 2, and mixed-precision (FP16 autocast) training for 15 epochs on a single GPU. Model selection used validation accuracy and loss tracked at the end of every epoch; the final-epoch weights were used for all reported results.

\section{Results and Discussion}
\label{sec:results}

\subsection{Training Dynamics}
Table~\ref{tab:training_curve} reports training and validation loss/accuracy at selected epochs. Training accuracy rises from 14.1\% at epoch 1 to 99.4\% at epoch 15, while validation accuracy rises more slowly and less monotonically, reaching 80.0\% at epoch 14 before settling at 77.5\% at epoch 15 — a pattern consistent with the model beginning to overfit the small training set in later epochs.

\begin{table}[t]
\centering
\caption{Training and validation loss/accuracy at selected epochs (15 total).}
\label{tab:training_curve}
\begin{tabular}{@{}lcccc@{}}
\toprule
Epoch & Train Loss & Train Acc. & Val. Loss & Val. Acc. \\
\midrule
1  & 2.415 & 0.141 & 2.108 & 0.250 \\
5  & 1.416 & 0.365 & 1.703 & 0.150 \\
8  & 0.953 & 0.724 & 1.237 & 0.475 \\
10 & 0.388 & 0.942 & 0.897 & 0.650 \\
12 & 0.174 & 0.955 & 0.637 & 0.775 \\
14 & 0.079 & 0.987 & 0.614 & 0.800 \\
15 & 0.083 & 0.994 & 0.681 & 0.775 \\
\bottomrule
\end{tabular}
\end{table}

\subsection{Validation Performance}
\label{sec:val_perf}
\begin{figure}[t]
    \centering
    \includegraphics[width=0.85\textwidth]{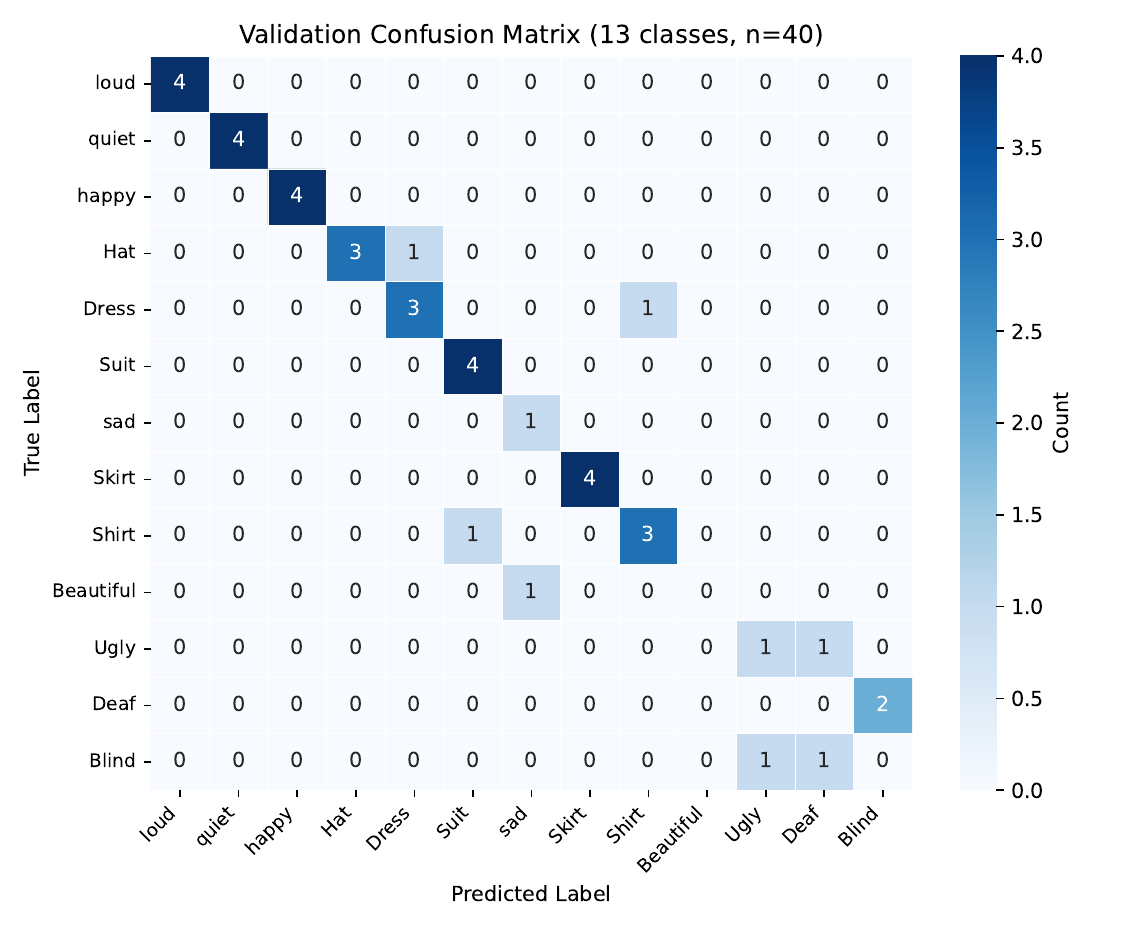}
    \caption{Validation confusion matrix across the 13 sign classes (40 clips total).}
    \label{fig:confmat}
\end{figure}

On the 40-clip validation set, the final model achieves an overall accuracy of 77.5\% (31/40 correct). Table~\ref{tab:classreport} reports per-class precision, recall, and F1, and Figure~\ref{fig:confmat} shows the corresponding confusion matrix.

\begin{table}[t]
\centering
\caption{Per-class precision, recall, and F1-score on the 40-clip validation set.}
\label{tab:classreport}
\begin{tabular}{@{}lcccc@{}}
\toprule
Class & Precision & Recall & F1 & Support \\
\midrule
loud      & 1.00 & 1.00 & 1.00 & 4 \\
quiet     & 1.00 & 1.00 & 1.00 & 4 \\
happy     & 1.00 & 1.00 & 1.00 & 4 \\
Hat       & 1.00 & 0.75 & 0.86 & 4 \\
Dress     & 0.75 & 0.75 & 0.75 & 4 \\
Suit      & 0.80 & 1.00 & 0.89 & 4 \\
sad       & 0.50 & 1.00 & 0.67 & 1 \\
Skirt     & 1.00 & 1.00 & 1.00 & 4 \\
Shirt     & 0.75 & 0.75 & 0.75 & 4 \\
Beautiful & 0.00 & 0.00 & 0.00 & 1 \\
Ugly      & 0.50 & 0.50 & 0.50 & 2 \\
Deaf      & 0.00 & 0.00 & 0.00 & 2 \\
Blind     & 0.00 & 0.00 & 0.00 & 2 \\
\midrule
\textbf{Accuracy}     & \multicolumn{4}{c}{\textbf{0.78 (40 samples)}} \\
Macro avg.    & 0.64 & 0.67 & 0.65 & 40 \\
Weighted avg. & 0.77 & 0.78 & 0.77 & 40 \\
\bottomrule
\end{tabular}
\end{table}

Five classes (\emph{loud}, \emph{quiet}, \emph{happy}, \emph{Suit}, \emph{Skirt}) are classified with perfect recall, and \emph{Hat}, \emph{Dress}, and \emph{Shirt} are recognized with 75\% recall. The model fails entirely (0\% recall) on \emph{Beautiful}, \emph{Deaf}, and \emph{Blind}, and performs at chance or below on \emph{Ugly}. We note that three of these four weakest classes (\emph{Beautiful}, \emph{Ugly}, \emph{Deaf}, \emph{Blind}) have the smallest validation support (1–2 samples each), since the underlying class-wise sample counts in this curated subset are themselves small and imbalanced; the reported precision/recall for these classes should therefore be read as indicative rather than statistically robust.

\subsection{Error Analysis}
\label{sec:error_analysis}
Inspecting the confusion matrix, two clusters of confusion stand out. First, a visually/posturally related cluster among the clothing-noun classes: \emph{Hat} is confused with \emph{Dress} (1 instance), \emph{Dress} is confused with \emph{Shirt} (1 instance), and \emph{Shirt} is confused with \emph{Suit} (1 instance). These four signs plausibly share overlapping hand placement near the upper torso and head, which may be difficult for the model to disambiguate from a 16-frame, $224\times224$ sample. Second, a semantic/affective cluster among \emph{Beautiful}, \emph{Ugly}, \emph{Deaf}, and \emph{Blind}: \emph{Beautiful} is entirely misclassified as \emph{sad}, and \emph{Ugly}/\emph{Deaf}/\emph{Blind} are mutually confused with each other in every error case. We hypothesize this reflects a combination of (i) genuinely overlapping facial-expression and hand-position cues across these signs, and (ii) the very small number of training examples per class limiting the model's ability to learn discriminative features for this subgroup specifically.

\subsection{Inference Pipeline}
\label{sec:inference}
We packaged the trained classifier and the NLLB translation model into an interactive Streamlit application, exposed during development via an ngrok tunnel for convenient browser access from the training notebook. A user uploads a short video (\texttt{.mp4}, \texttt{.mov}, etc.); the app extracts 16 uniformly spaced frames with OpenCV, applies the same resize/normalize transform used in training, and runs a forward pass through the fine-tuned VideoMAE model to obtain a predicted label and confidence score. The cleaned label is then translated independently into Hindi, Telugu, and Bengali via three separate NLLB-200 calls (one per target language code), and all four outputs (English label, three translations) are rendered to the user. In a representative session, a held-out clip of the sign for ``happy'' was classified correctly with 0.92 confidence; the predicted label was translated to the Hindi, Telugu, and Bengali words for ``happy'' in their respective native scripts (Devanagari, Telugu, and Bengali script), which are reproduced in the rendered demo output but are omitted here due to font limitations in the manuscript build.

\subsection{Discussion of Findings}
The results show that a VideoMAE encoder pretrained on general video data can be fine-tuned, with a modest number of labeled clips per class, to recognize isolated sign-language gestures with reasonable validation accuracy (77.5\% over 13 classes, well above the 7.7\% random baseline). The gap between training accuracy ($>$99\%) and validation accuracy ($\sim$78\%) indicates overfitting given the small per-class sample sizes, which is unsurprising given that several classes have only 1–4 validation examples. The error patterns in Section~\ref{sec:error_analysis} suggest that scaling up the number of clips per class, and adding more signers per class to increase intra-class style diversity, would likely improve both raw accuracy and the reliability of the per-class metrics. On the translation side, because the second stage translates isolated English words rather than full sentences, the NLLB model has no sentence-level context to disambiguate polysemous words (e.g., ``suit'' as clothing vs.\ as in ``to suit someone''); this is a structural limitation of word-level translation rather than a deficiency of the NLLB model itself, and is discussed further below.

\section{Limitations}
\label{sec:limitations}

\begin{enumerate}[leftmargin=1.4em]
    \item \textbf{Small, imbalanced label set.} Compute and storage constraints limited training to 13 classes drawn from a much larger corpus, with as few as 1–2 validation clips for several classes; reported per-class metrics for these classes are necessarily noisy.
    \item \textbf{Isolated-word, not continuous, signing.} The system classifies pre-segmented clips into a fixed vocabulary; it does not parse continuous signed sentences or generate free-form English text, and cannot recognize signs outside its trained label set.
    \item \textbf{Signer and recording variability.} Differences in lighting, background, camera angle, frame rate, and individual signing style are not explicitly modeled, and may degrade performance on signers or recording conditions unlike those in the training data.
    \item \textbf{Single-word translation ambiguity.} Translating isolated words without sentence context can produce ambiguous or context-inappropriate translations for polysemous English words.
    \item \textbf{Offline, non-real-time operation.} The current implementation targets batch/offline inference through a Streamlit demo; real-time or streaming video translation is out of scope for this phase of work.
\end{enumerate}

\section{Conclusion and Future Work}
\label{sec:conclusion}

We presented a two-stage pipeline for translating Indian Sign Language video gestures into English and subsequently into Hindi, Telugu, and Bengali, combining a fine-tuned VideoMAE video transformer for gesture classification with a pretrained NLLB-200 model for multilingual translation. On a 13-class subset of the AI4Bharat ISL corpus, the approach reaches 99\% training accuracy and 78\% validation accuracy, with most classification errors concentrated among visually or semantically similar sign pairs. We view this as an initial, small-scale feasibility study rather than a production-ready system, and we are releasing the training and inference code to support reproducibility and extension by the community.

Future work includes: (i) scaling to the full AI4Bharat vocabulary and additional ISL corpora to test generalization across a much larger label set; (ii) moving from isolated-word classification toward continuous sign language translation that can generate full English sentences from unsegmented video, e.g., using sequence-to-sequence or CTC-based decoding over the VideoMAE feature stream; (iii) extending multilingual support to additional Indian languages within the NLLB-200 language set; and (iv) optimizing model size and inference latency (e.g., via distillation or quantization) for deployment on edge or mobile devices to support real-time use in classrooms and public services.

\section*{Acknowledgements}
We thank the Department of Computer Science and Engineering, IIT Patna, for providing the research environment for this work, and the creators and curators of the AI4Bharat Indian Sign Language dataset at IIT Madras, without which this study would not have been possible. We also acknowledge the open-source contributions of the VideoMAE and NLLB development teams.

\section*{Code Availability}
Code for data loading, model fine-tuning, evaluation, and the Streamlit inference demo is available at: \url{https://github.com/rnandip/Deep-Learning-Projects}.

\bibliographystyle{unsrtnat}
\bibliography{references}

\end{document}